\documentclass[10pt,twocolumn,letterpaper]{article}

\usepackage{cvpr}              

\definecolor{cvprblue}{rgb}{0.21,0.49,0.74}
\usepackage[pagebackref,breaklinks,colorlinks,allcolors=cvprblue]{hyperref}

\def\paperID{5350} 
\def\confName{CVPR}
\def\confYear{2025}

\title{Associative Transformer}

\newcommand*\samethanks[1][\value{footnote}]{\footnotemark[#1]}
\author{\normalsize Yuwei Sun\textsuperscript{1,2}, Hideya Ochiai\textsuperscript{3}, Zhirong Wu\textsuperscript{4}, Stephen Lin\textsuperscript{4\thanks{Both authors contributed equally to the supervision of this work.\\
Corresponding author: Yuwei Sun (yuwei\_sun@araya.org).}}, Ryota Kanai\textsuperscript{1\samethanks}\\
\normalsize \textsuperscript{1}Araya Research, \textsuperscript{2}RIKEN AIP, \textsuperscript{3}The University of Tokyo, \textsuperscript{4}Microsoft Research
}

\usepackage{hyperref}
\usepackage{url}
\usepackage{comment}
\usepackage{amsmath,amsfonts,amssymb}
\usepackage{algorithmic}
\usepackage{algorithm}
\usepackage{array}
\usepackage{caption}
\usepackage{subcaption}
\usepackage{textcomp}
\usepackage{stfloats}
\usepackage{verbatim}
\usepackage{graphicx}
\usepackage{comment}
\usepackage{color}
\usepackage{bbm}
\usepackage{multirow}
\usepackage{wrapfig}
\usepackage{enumitem}

\begin{document}
\maketitle

\begin{abstract}
  Emerging from the pairwise attention in conventional Transformers, there is a growing interest in sparse attention mechanisms that align more closely with localized, contextual learning in the biological brain. Existing studies such as the Coordination method employ iterative cross-attention mechanisms with a bottleneck to enable the sparse association of inputs. However, these methods are parameter inefficient and fail in more complex relational reasoning tasks. To this end, we propose \textbf{A}ssoc\textbf{i}ative \textbf{T}ransformer (AiT) to enhance the association among sparsely attended input tokens, improving parameter efficiency and performance in various vision tasks such as classification and relational reasoning. AiT leverages a learnable explicit memory comprising specialized priors that guide bottleneck attentions to facilitate the extraction of diverse localized tokens. Moreover, AiT employs an associative memory-based token reconstruction using a Hopfield energy function. The extensive empirical experiments demonstrate that AiT requires significantly fewer parameters and attention layers outperforming a broad range of sparse Transformer models. Additionally, AiT outperforms the SOTA sparse Transformer models including the Coordination method on the Sort-of-CLEVR dataset.
\end{abstract}

\section{Introduction}

Transformer models use pairwise attention to establish correlations among disparate segments of input information \cite{attention,vit}. Emerging from the pair-wise attention, there is a growing interest in leveraging sparse interactions for localized, contextual learning. This sparsity attribute has demonstrated advantages in enhancing learning performance and efficiency \cite{brook91, binding}. Sparse knowledge association can find resonance with the neuroscientific grounding of the Global Workspace Theory (GWT) \cite{Baars1988,dehaene98,vanr20,juliani2022}. GWT explains a fundamental cognitive architecture for working memory where diverse specialized modules compete to write information into a shared workspace through a communication bottleneck. The bottleneck facilitates the processing of content-addressable information using attention guided by contents in the shared workspace \cite{cog2,cog1}.

A bottleneck guides models to generalize in a manner consistent with the underlying data distribution through inductive biases of sparsity \cite{nfl1,nfl2}, resulting in superior performance. In this regard, the Coordination method \cite{goyal} is an initial attempt to assess the effectiveness of such a bottleneck in Transformers. Unfortunately, its design is parameter inefficient and often fails in more complex relational reasoning tasks \cite{sort,shanahan2020explicitly,mittal2021compositional,spies2022sparse,kerg2022neural,jiang2023object,mondal2024slot}. In this work, we aim to enhance sparse attention capability of Transformers by leveraging low-rank explicit memory to learn a diverse set of priors to guide the attention.
Furthermore, drawing inspiration from associative memory based on Hebbian learning, we utilize continuous Hopfield networks \cite{johnhf,hfl} to reconstruct input tokens from the explicit memory. The associative memory enhances the effective association of sparsely attended historical inputs. 

To this end, we propose \textbf{A}ssoc\textbf{i}ative \textbf{T}ransformer (AiT), a sparse representation learner. AiT learns explicit memory for each attention layer to facilitate the extraction of localized features at different abstraction levels. The explicit memory evolves into specialized priors through the bottleneck attention, with each focusing on a specific spatial relation of input tokens. We further propose a bottleneck attention balance loss to encourage diversity among the learned priors. Furthermore, the acquired priors function as distinct attractors within the associative memory of Hopfield networks, facilitating token reconstruction from memory. Extensive empirical results demonstrate that the emerging specialization of priors and the reconstruction within associative memory in AiT could significantly enhance parameter efficiency and model performance across a broad spectrum of tasks. 

Overall, our main contributions are three-fold:

1) We propose the Associative Transformer (AiT) to tackle inefficient sparse attention mechanisms in conventional Transformers, leveraging emerging specialized priors for guided bottleneck attention and token reconstruction within associative memory (Section \ref{sec:bottleneck}).

2) The learned priors function as attractors of Hopfield networks, facilitating information correlation and retrieval from historical inputs. To the best of our knowledge, this is the first study on consolidating Hopfield networks with the sparse attention mechanism (Section \ref{sec:information}).

3) Extensive experiments including an ablation study demonstrate enhanced parameter efficiency of AiT in various classification and relational reasoning tasks, outperforming the Coordination method (Section \ref{sec:exp}).


\section{Related work}

This section provides a summary of the most relevant recent work on sparse attention, latent memory-enabled Transformers, and external memory mechanisms. We investigate the relatedness of these studies to various properties of the Global Workspace Theory (GWT) in Appendix \ref{appe:rela}.

\paragraph{Sparse Transformers}
Studies on sparse Transformers have explored consolidating latent memory to extract localized, contextual representations from inputs \cite{lee19,gmat,luna,jaegle21,goyal,jaegle22}. For example, Perceiver \cite{jaegle21,jaegle22} utilized iterative cross-attention with a latent array and transformation to capture dependencies across input tokens. Set Transformer \cite{lee19} and Linear Unified Nested Attention \cite{luna} also employed iterative cross-attention but without the latent transformation. Other methods, like Blockwise Self-Attention, relied on a strong assumption of predefined modularization for attention specialization \cite{qiu20}. In contrast, our method learns such modularization and attention specialization through end-to-end training. Unlike the vast majority of methods that employ latent memory, our proposed method leverages both explicit memory and associative memory for enhanced contextual learning capability.

In a similar approach to constructing sparse Transformers based on GWT, the Coordination method \cite{goyal} introduced a bottleneck into cross-attention mechanisms to encourage neural module specialization. The results indicated that the competition via the bottleneck could contribute to a small set of specialized priors in latent memory to facilitate relational learning. However, the number of priors was limited to fewer than 10, all with the same dimension as the input representation. By contrary, our method employs low-rank explicit memory to learn a significantly richer set of priors (up to 128 priors learned from a pool of 32.8k tokens). Moreover, the Coordination method relies on iterative cross-attentions, while our work focuses on novel associative memory-enabled sparse attention.

This work is also related to modular neural networks and mixture of experts \cite{modular3,modular4} in terms of competition in the shared workspace. Separating the information processing within Transformers into distinct components, depending on the input, has shown more flexibility in data processing.

\paragraph{Memory mechanisms}
External memory such as tape storage and associative memory, has been employed in deep neural networks \cite{turing,modernhf,chan18}, with recent studies exploring the potential use of Hopfield networks and their modern variants \cite{johnhf,demi,hfl}. In this work, we incorporate Hopfield networks as an integral element in the sparse attention mechanism of the proposed Associative Transformer. This approach is fundamentally different from the previous studies such as Energy Transformer \cite{hoover23} that mainly focused on utilizing Hopfield networks independently of the attention mechanism.

\section{Associative Transformer}
\label{sec:gwl}

\begin{figure*}[t]
    \centering
    \includegraphics[width=\linewidth]{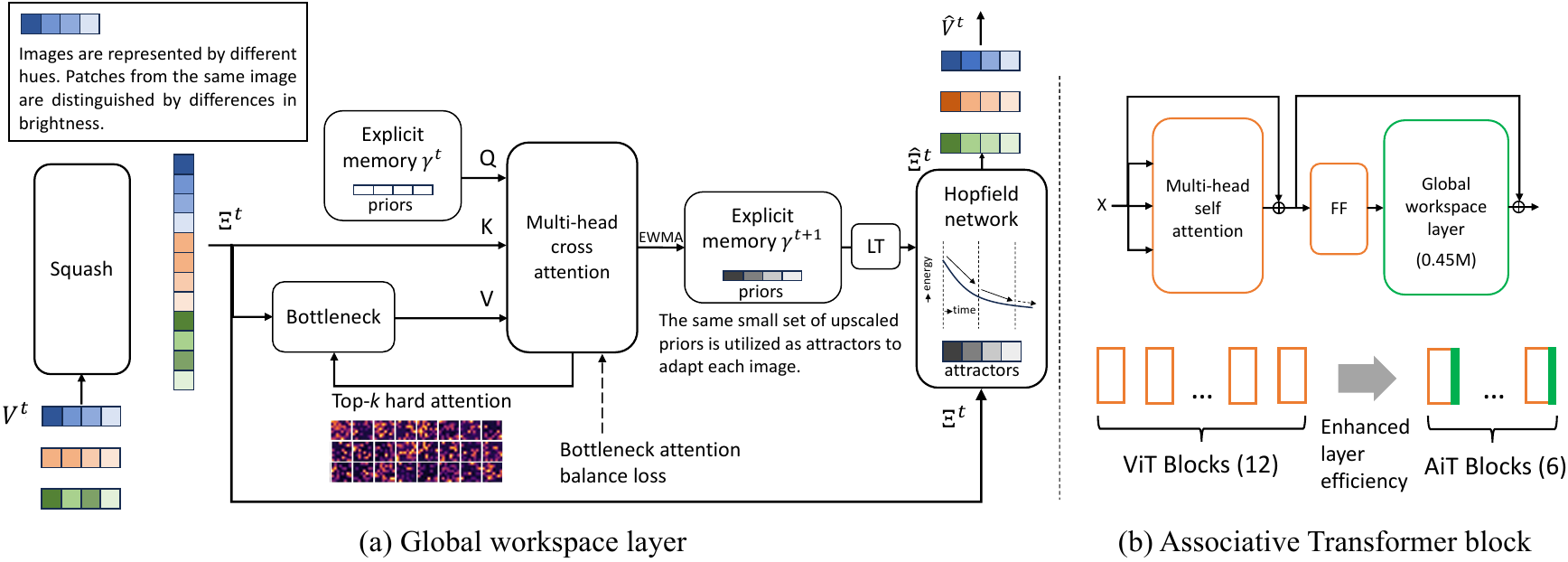}
    \caption{The scheme of the Associative Transformer (AiT). (a) In a global workspace layer, the input $\mathbb{R}^{B\times N\times E}$ is squashed into vectors $\mathbb{R}^{BN\times E}$. The squashed representations are projected to a latent space of dimension $D<<E$ and are sparsely selected to update the explicit memory via a fixed bottleneck $k<<BN$. The Hopfield network utilizes the memory to reconstruct the input tokens, where a learnable linear transformation (LT) scales the memory contents back to the input dimension $E$. (b) The Associative Transformer block consists of self attention, feed-forward layers, and the global workspace layer. Compared to Vision Transformer (ViT), leveraging the global workspace layer enhances the layer efficiency. A shallower 6-layer AiT is shown to outperform a 12-layer ViT (see Table \ref{tab:class}).}
    \label{fig:sche}
    
\end{figure*}


This section discusses the key technical underpinnings of the Associative Transformer (AiT), focusing on the newly devised \textit{Global Workspace Layer (GWL)}. The GWL integrates three main elements: low-rank explicit memory, bottleneck attention, and associative memory (Figure \ref{fig:sche}). It facilitates efficient writing and reading in explicit memory by promoting competition among input tokens through a specialized top-$k$ bottleneck attention mechanism that limits cross-attention capacity. Two techniques are employed to maintain balanced information writing: Exponential Weighted Moving Average (EWMA), which stabilizes memory updates by combining old and new information, and a novel bottleneck attention balance loss, which encourages diverse token selection and counteracts the loss of specificity across layers. After writing, the explicit memory contents are broadcast back to the input tokens for retrieval within a Hopfield network-based associative memory, reconstructing the tokens into more globally meaningful representations that serve as the output of the GWL. In practice, the GWL can be integrated with conventional attention layers to enhance the efficiency of Transformer models, boosting their ability to learn abstract spatial relations across samples and time steps.

\subsection{Vision Transformers for classification tasks}

Vision Transformers (ViT) tackle image classification tasks by processing sequences of image patches. The pre-processing layer partitions an image into non-overlapping patches. Let $x\in\mathbb{R}^{H\times W\times C}$ be an input from the dataset $X$, where $(H, W)$ is the resolution of the image and $C$ is the number of channels. $x$ is separated into a sequence of patches $x_p\in \mathbb{R}^{N\times (P^2\cdot C)}$, where $(P,P)$ is the resolution of each image patch and $N=\frac{HW}{P^2}$ is the number of patches. These patches are mapped to embeddings $v_p\in\mathbb{R}^{N\times E}$ with the linear projection. ViT leverages self-attention to map a query and a set of key-value pairs to an output. The patch embeddings are used to obtain the query, key, and value based on linear transformations $W^Q \in \mathbb{R}^{E\times D},\,W^K \in \mathbb{R}^{E\times D}, \text{ and, } W^V \in \mathbb{R}^{E\times D}$. The output of an individual attention head is a weighted sum of the values, $\mbox{softmax}(\frac{W^Q_iv(W^K_iv)^T}{\sqrt{D}})\,W^V_iv$.

We specifically consider supervised learning with $F$ categories. Let $y \in \{1,2,...,F\} = Y$ denote a label. Given a Transformer model $f(x)$, the prediction is given by $\hat{y} = \text{arg}\max_j f(x)_j$ where $f(x)_j$ denotes the $j$th element of $f(x)$. The training is attained by minimizing the loss with respect to the model parameter $\theta$:
\begin{equation}
\ell_\text{class}(\theta)= J(Y, f(X; \theta)),
\end{equation}
where $J$ denotes the cross entropy loss.

\subsection{Low-rank priors in explicit memory}

To induce memory in large-scale models like Transformers, we focus on learning more efficient explicit memory, which is later consolidated with Hopfield networks. This allows us to decouple the memory storage from the Hopfield network learning, such that the memory could be gradually updated with old and new memory using methods such as exponentially weighted moving average (EWMA). The explicit memory is only trainable during model training and frozen during evaluation. The proposed Global Workspace Layer enables efficient writing and reading in the explicit memory to learn a set of low-rank priors through a tailored bottleneck attention mechanism.

In particular, we employ a squash layer to concatenate patch representations within the entire batch $V\in\mathbb{R}^{B\times N\times E}$ into vectors $\Xi\in\mathbb{R}^{BN\times E}$, fostering association among tokens across samples. In practice, the number of tokens $B \times N$ varies depending on the batch size $B$. The attention bottleneck with a fixed capacity $k$ restricts the number of attended tokens to a constant value. The output of the squash layer is then attended to by the multi-head cross-attention using the contents of the explicit memory as queries. The explicit memory is comprised of $M$ learnable low-rank priors $\gamma = \mathbb{R}^{M\times D}$ where $D$ is the dimension of the prior, $D << E$. Priors are general assumptions about tokens to guide the attention for the token selection. The priors are utilized as various queries to compute cross-attentions that extract various sets of tokens and update the explicit memory. We found that a rank of 32 and initialization of the priors with a Gaussian distribution works well in most cases. 

We devise a tailored cross-attention mechanism to update the explicit memory using the squash layer's output $\Xi \in\mathbb{R}^{BN\times E}$. Notably, in the cross-attention, the query is a function of the current memory content $\gamma = \{\gamma_i\}_{i=1}^M$. The key and value are functions of the squash layer's output $\Xi$. The attention scores $A_i$ for head $i$ are computed as follows $\text{A}_i(\gamma, \Xi)=\text{softmax}(\frac{\gamma(\Xi W^K_{i})^T}{\sqrt{D}})$. This is the case of soft attention without constraints on the bottleneck capacity.

\subsection{Bottleneck attention with a limited capacity}
\label{sec:bottleneck}

We discuss the case of limiting the cross-attention capacity via top-$k$ bottleneck attention. The bottleneck attention constrains the number of tokens that can be attended to, fostering competition among tokens and ensuring the selection of the most relevant tokens. Notably, we select the tokens with the top-$k$ highest attention scores from $A_i$, ${h}_i = \text{top-}k(\text{A}_i)\Xi W^V$. Additionally, to ensure a stable update of the explicit memory across different time steps, we employ layer normalization and the exponentially weighted moving average (EWMA) strategies as follows:
\begin{equation}
\label{eq:attention}
\begin{aligned}
\hat{\gamma}^{t+1} = (1 - \alpha) \cdot \gamma^{t} + \alpha \cdot \text{LN}(\text{Concat}({h}_1,\dots,{h}_S)W^O),\\
\gamma^{t+1} = \frac{{\hat{\gamma}^{t+1}}}{{\sqrt{\sum_{{j=1}}^{M} (\hat{\gamma}^{t+1}_j)^2}}},
\end{aligned}
\end{equation}
where $t$ is the batch time step, LN is the layer normalization, $S$ is the number of attention heads, and $\alpha$ is a smoothing factor determining the decay rate of older observations, which is a small value such as 0.1. EWMA ensures stable memory updates by accumulating old and new memories. We found that different batch sizes $B$ have little influence on the test performance. This can be attributed to the small decay rate that results in the stable memory update.

\paragraph{Bottleneck attention balance loss}

The sparsity induced by the bottleneck attention results in the emergence of specialized priors by attending to various tokens. However, cascading multiple bottleneck attentions across Transformer layers could lead to difficulty in efficiently forming specialized priors. As information flows through multiple bottleneck attention layers, the representations become diluted, necessitating a mechanism to counteract this inherent loss of input specificity for the effective selection of meaningful tokens across layers. To overcome this challenge, we propose the \textit{bottleneck attention balance loss} to encourage a more diverse token selection with the bottleneck attention. The bottleneck attention balance loss $\ell_{\text{bottleneck}}$ comprises two components, i.e., the cumulative attention loss $\ell_{\text{{importance}}}$, which computes the total assigned attention scores for each token, and the selected instance loss $\ell_{\text{{loads}}}$, which provides an estimate of the frequency of each token being selected. Then, the loss is computed for each token position $o \in \{1,2,\dots,B \times N\}$ in the squash layer output. We devise the bottleneck attention balance loss for attention head $i$ by measuring the normalized variances of the cumulative attention loss and the selected instance loss across token positions as follows:
\begin{equation}
    \ell_{\text{importance}_{i,o}}  = \sum_{j=1}^M \text{A}_{i,j,o},\,
    \ell_{\text{loads}_{i,o}}  = \sum_{j=1}^M (\text{A}_{i,j,o} > 0),
\end{equation}
\begin{equation}
\begin{aligned}
    \ell_{\text{bottleneck}_{i}}  = \frac{{\text{Var}(\{\ell_{\text{importance}_{i,o}}}\}_{o=1}^{B \times N})}{{(\frac{1}{B \times N}\sum_{o=1}^{B \times N}\ell_{\text{importance}_{i,o}}})^2 + \epsilon} + \\ \frac{{\text{Var}(\{\ell_{\text{loads}_{i,o}}}\}_{o=1}^{B \times N})}{{(\frac{1}{B \times N}\sum_{o=1}^{B \times N}\ell_{\text{loads}_{i,o}}})^2 + \epsilon},
\end{aligned}
\end{equation}
where $\text{A}_{i,j,o}$ is head $i$'s attention score for the $j$th prior at token position $o$, Var($\cdot$) denotes the variance, and $\epsilon$ is a small value to avoid division by zero. 

The losses for the different heads are summed up and added to the classification loss,
$\ell = \ell_{\text{class}} + \sigma \cdot \sum_{i=1}^S \ell_{\text{bottleneck}_i},$
where $\sigma$ is a coefficient to adjust the ratio between the classification loss and the bottleneck attention balance loss.

\subsection{Information retrieval with associative memory}
\label{sec:information}

After writing information from the input tokens into the explicit memory, the learned priors serve as attractors for a novel information retrieval component based on associative memory. The aim is to reconstruct the current input tokens towards more globally meaningful representations learned and stored in the explicit memory for enhanced token representations. The proposed architecture is an attractor network where a token converges to one of these attractors derived from the priors stored in the explicit memory, by decreasing its energy with respect to the explicit memory.

\paragraph{Attractors}

The learned priors in the explicit memory function as attractors within associative memory of a Hopfield network. Attractors have basins of attraction, and any input that enters an attractor’s basin converges to that specific attractor. Attractors usually have the same dimension as input states. Since we employ low-rank priors $\gamma^{t+1} \in \mathbb{R}^{M\times D}$, a learnable linear transformation $f_{\text{LT}}(\cdot)$ projects the priors into the same dimension $E$ as the input tokens before utilizing them as attractors.

\paragraph{Energy-based retrieval}

The upscaled priors $f_{\text{LT}}(\gamma^{t+1})$ are stored within the associative memory of a continuous Hopfield network \cite{johnhf,hfl} as various attractors to reconstruct the input tokens $\Xi^t$. In particular, we employ the Hopfield energy function to evolve the input tokens into more globally meaningful representations with respect to these attractors. The attractors $f_{\text{LT}}(\gamma^{t+1})$ are shared across instances of the input tokens $\xi^t\in\Xi^t$.  We reconstruct the input tokens using the continuous Hopfield network and the learned attractors in explicit memory as follows:
\begin{equation}
\begin{aligned}
\label{eq:ener}
    E(\Xi^t) = -\text{lse}(\beta,f_{\text{LT}}(\gamma^{t+1}) \Xi^t)+\frac{1}{2}\Xi^t{\Xi^t}^T+ \\
    \beta^{-1}\text{log}M+\frac{1}{2}(\max_{i} |f_{\text{LT}}(\gamma^{t+1}_i)|)^2,\\
\end{aligned}
\end{equation}
\begin{equation}
    \hat{\Xi}^t = \arg\min_{\Xi^t} E(\Xi^t),
\end{equation}
where $t$ is the batch time step, $\beta$ is an inverse temperature variable, lse is the log-sum-exp function, and $\underset{i}{\text{max}} |f_{\text{LT}}(\gamma^{t+1}_i)|$ represents the largest norm of attractors. We employ an iterative update procedure for a fixed number of steps. All tokens reach their minimum at the same time in the associative memory with their global energy decreasing. For training efficiency, we employed a single step to reconstruct the tokens as suggested in \cite{hfl}. Thus, the reconstructed tokens $\hat{\Xi}^t \in \mathbb{R}^{(B \times N) \times E}$ are transformed into $\hat{V}^t \in \mathbb{R}^{B \times N \times E}$ as the output of the global workspace layer. Moreover, depending on $\beta$, the reconstructed token $\hat{\Xi}^{t}$ can be either metastable states representing mixtures of attractors or fixed states represented by one of the attractors. We discuss the tuning of $\beta$ in Appendix \ref{appe:hopf} and demonstrate the complete algorithm in Appendix \ref{appe:algo}.

\section{Experiments}
\label{sec:exp}

In this section, we present the settings and extensive experiment results for classification and relational reasoning tasks. We compare AiT with a broad range of conventional sparse Transformers in terms of model performance and parameter efficiency. Comprehensive ablation studies demonstrate the effectiveness of the bottleneck attention and associative memory for enhanced model performance.

\subsection{Settings}

\paragraph{Datasets}
We evaluate model performance based on three different types of datasets: (1) small classification tasks (Triangle \cite{goyal}, CIFAR10, and CIFAR100 \cite{cifar100}), (2) medium and large-sized classification tasks (Oxford Pet \cite{pet} and ImageNet100 \cite{deng2009imagenet}), and (3) relational reasoning tasks (Sort-of-CLEVR \cite{sort}). We train the model on these datasets from scratch using the training split and evaluate using the test split. A detailed description of the datasets can be found in Appendix \ref{appe:data}.

\paragraph{Model variants}

We investigate three different sizes of model configurations, i.e., Small, Medium, and Base. The Base variant setting employs 12 layers, 12 attention heads for each layer, a hidden dimension of 768, and an MLP dimension of 3072. The Medium variant has 6 layers, and the Small variant has 2 layers. In what follows, we use brief notation to indicate the model size, for instance, AiT-B means the “Base” variant of the Associative Transformer. Note that the added Global Workspace Layer is light-weight using 0.45M parameters. 

\paragraph{Hyperparameters}

The hyperparameters were chosen based on a grid search. A batch size of 512 was employed for the CIFAR datasets and the Triangle dataset, 128 for the Pet dataset, and 64 for the Sort-of-CLEVR dataset. We utilized the AdamW optimizer with $\beta_1=0.9$, $\beta_2=0.999$, and a weight decay of 0.01. A cosine learning rate scheduler was implemented with an initial learning rate of 1e-5, a warm-up phase of 5 (15) epochs within a total of 100 (300) epochs, and a minimum learning rate set to 1e-6. The smoothing factor of the exponentially weighted moving average, the coefficient $\sigma$, and the small value $\epsilon$ in the bottleneck balance loss were set to 0.1, 1e-2, and 1e-10, respectively. 
We employed a bottleneck size of 512 for CIFAR and Pet, 64 for Triangle, and 256 for Relational Reasoning. We utilized 32 memory slots for CIFAR, Triangle, and Relational Reasoning, and 128 slots for Pet. Unless otherwise noted, we trained the model for 100 epochs and reported the mean of three individual experiments. All experiments were based on PyTorch and four A100 GPUs, and the code would be made publicly available. The detailed experimental settings and hyperparameters are presented in Appendix \ref{appe:hype}.

\subsection{Classification tasks}
\label{sec:classification}

\begin{table*}[t]
    \small
    \centering
    \renewcommand{\arraystretch}{1.12}
    \begin{tabular}{l|cccccc}
    \toprule
        Methods & CIFAR10 (\%) & CIFAR100 (\%) & Triangle (\%) & Average (\%) & Size (M) & \#FLOPs \\ \hline
        AiT-Base & \textbf{85.44 $\pm$ 0.31} & \textbf{60.78 $\pm$ 0.25} & \textbf{99.64 $\pm$ 0.14} & \textbf{81.95} & 91.0 & 5.77$\times 10^9$ \\ 
        AiT-Medium & 84.59 $\pm$ 0.27 & 60.58 $\pm$ 0.32 & 99.57 $\pm$ 0.16 & 81.58 & 45.9 & 2.89$\times 10^9$ \\ 
        AiT-Small & 83.34 $\pm$ 0.44 & 56.30 $\pm$ 0.38 & 99.47 $\pm$ 0.09 & 79.70 & 15.8 & 9.64$\times 10^8$ \\  \hline
        Coordination \cite{goyal} & 75.31 $\pm$ 0.72 & 43.90 $\pm$ 0.21 & 91.66 $\pm$ 0.56 & 70.29 & 2.2 & 1.46$\times 10^8$ \\
        Coordination-DH \cite{goyal} & 72.49 $\pm$ 0.61 & 51.70 $\pm$ 0.75 & 81.78 $\pm$ 0.59 & 68.66 & 16.6 & 3.15$\times 10^8$ \\
        Coordination-D \cite{goyal} & 74.50 $\pm$ 0.68 & 40.69 $\pm$ 0.39 & 86.28 $\pm$ 0.73 & 67.16 & 2.2 & 2.91$\times 10^8$ \\
        Coordination-H \cite{goyal} & 78.51 $\pm$ 0.58 & 48.59 $\pm$ 0.72 & 72.53 $\pm$ 0.35 & 66.54 & 8.4 & 1.59$\times 10^8$ \\ \hline
        ViT-Base \cite{vit} & 83.82 $\pm$ 0.17 & 57.92 $\pm$ 0.40 & 99.63 $\pm$ 0.15 & 80.46 & 85.7 & 5.60$\times 10^9$ \\ 
        ViT-Medium \cite{vit} & 82.41 $\pm$ 0.11 & 55.78 $\pm$ 0.09 & 99.62 $\pm$ 0.04 & 79.27 & 42.7 & 2.81$\times 10^9$ \\
        ViT-Small \cite{vit} & 79.53 $\pm$ 0.36 & 53.19 $\pm$ 0.37 & 99.47 $\pm$ 0.07 & 77.40 & 14.9 & 9.36$\times 10^8$ \\
        Perceiver \cite{jaegle21} & 82.52 $\pm$ 0.82 & 52.64 $\pm$ 0.44 & 96.78 $\pm$ 0.32 & 77.31 & 44.9 & 2.37$\times 10^9$ \\ 
        Set Transformer \cite{lee19} & 73.42 $\pm$ 0.43 & 40.19 $\pm$ 0.53 & 60.31 $\pm$ 0.29 & 57.97 & 2.2 & 1.11$\times 10^8$ \\
        BRIMs \cite{mittal20} & 60.10 $\pm$ 0.50 & 31.75 $\pm$ 0.28 & 58.34 $\pm$ 0.43 & 50.06 & 4.4 & 1.43$\times 10^8$ \\ 
        Luna \cite{luna} & 47.86 $\pm$ 0.53 & 23.38 $\pm$ 0.06 & 57.26 $\pm$ 0.19 & 42.83 & 77.6 & 5.08$\times 10^9$ \\
        \bottomrule
    \end{tabular}
    \caption{Performance comparison in the classification tasks.}
    \label{tab:class}
\end{table*}

\begin{table}[!h]
\centering
\small
\begin{tabular}{l|cc}
\toprule
Methods & Test Accuracy (\%) & Size (M)\\ \hline
AiT-Medium & \textbf{36.72} & 45.9\\ 
AiT-Small & 33.84 & 15.8\\ 
ViT-Base &  34.62 & 85.7\\ 
ViT-Medium & 31.72 & 42.7\\ 
ViT-Small & 28.16 & 14.9\\ 
\bottomrule
\end{tabular}
\caption{Performance comparison between AiT and ViT on the ImageNet100 dataset.}
\label{tab:imagenet100}
\end{table}

The experiments on image classification tasks include comparisons to a broad range of methods. We employed the author-recommended hyperparameters to re-implement these methods. In particular, for the Coordination method \cite{goyal}, we investigated the various model configurations: (1) Coordination consists of 4 layers with parameter sharing among different attention layers, (2) Coordination-D is a deeper model with 8 layers using parameter sharing, (3) Coordination-H is a high memory model with 4 layers that employ individual parameters, and (4) Coordination-DH is a high memory model with 8 layers. 

Table \ref{tab:class} demonstrates that AiT performance increased when scaling it from Small to Base, while the Coordination method could not scale with the increasing parameter sizes. AiT achieved better performance compared to Vision Transformers and the other sparse Transformers including Perceiver, Set Transformer, BRIMs, and Luna. Moreover, AiT relies on significantly fewer layers and parameters while outperforming Vision Transformers. Compared to ViT-Base using 12 layers and 85.7M parameters, AiT-Medium has only 6 layers and 45.9M parameters. Nevertheless, AiT-Medium exhibited an average performance of 81.58\%, surpassing ViT-Base's performance of 80.46\%. Table \ref{tab:imagenet100} shows additional comparison between AiT and ViT on the ImageNet100 dataset. AiT achieved superior performance in the various classification tasks using a much smaller parameter size.

Furthermore, compared to ViT-Base with 5.77$\times 10^9$ FLOPs during training, AiT-Medium utilized only 2.89$\times 10^9$ FLOPs. We compared with the other baseline methods, where the Coordination does not scale well with more model complexity and FLOPs, resulting in degraded performance. In contrast, the proposed Global Workspace Layer is a lightweight architecture, greatly enhancing the classification task performance and parameter efficiency of the conventional sparse Transformers.

\begin{figure}
    \centering
    \includegraphics[width=0.95\linewidth]{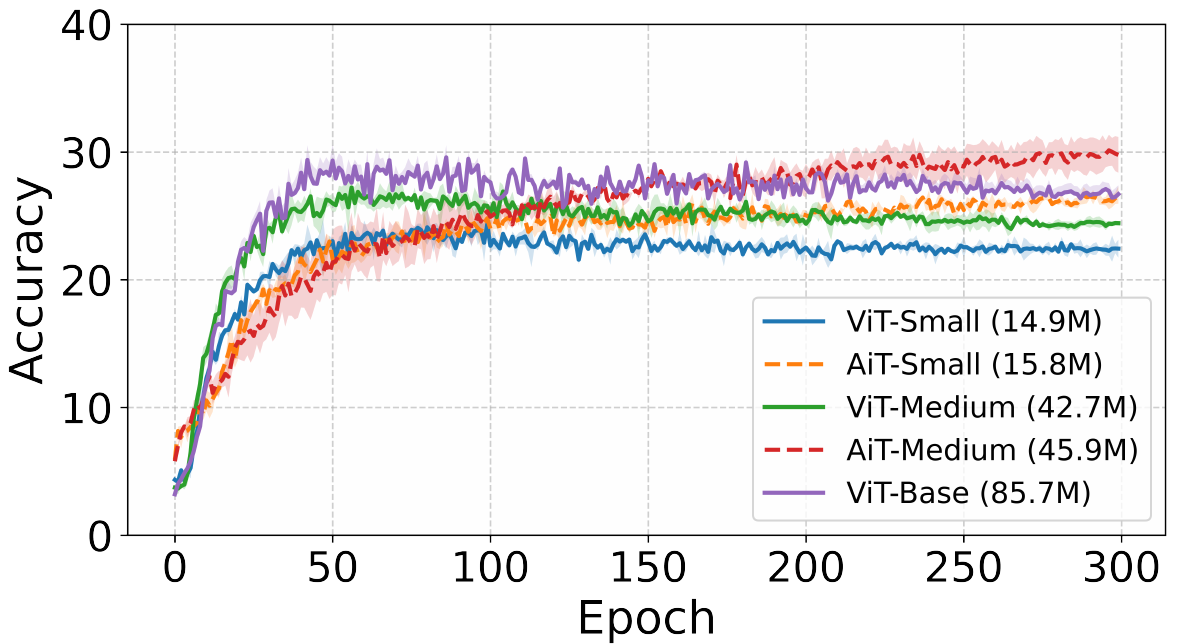}
    \caption{Comparison on the Pet dataset. AiT-Medium demonstrated a stable increase in performance outperforming the ViT-Base.}
    \label{fig:pet}
\end{figure}

We extended the evaluation to the Oxford Pet dataset \cite{pet}. We trained the model from scratch for 300 epochs. Figure \ref{fig:pet} demonstrates that AiT-Small with only additional 0.9M parameters (two Global Workspace Layers) increased ViT-Small's performance by 13\%. Although ViT-Base is a much larger model with 85.7M parameters with the optimized model regularizations, its accuracy quickly dropped after 50 epochs. In contrast, AiT-Medium with 45.9M parameters demonstrated a stable increase in performance and outperformed the ViT-Base model.

\subsection{Ablation study}
\label{sec:ablation}

\begin{table*}[t] 
\small
\centering
\renewcommand{\arraystretch}{1.12}
\begin{tabular}{l|cccc}
\toprule
Models & CIFAR10 (\%) & CIFAR100 (\%) & Triangle (\%) & Average (\%)\\ \hline
AiT & \textbf{83.34 $\pm$ 0.44} & \textbf{56.30 $\pm$ 0.38} & \textbf{99.47 $\pm$ 0.09} & \textbf{79.70}\\ \hline
Reset Memory & 81.94 $\pm$ 0.41 & 55.96 $\pm$ 0.39 & 99.46 $\pm$ 0.04 & 79.12\\
W/O Attention Balance Loss & 81.89 $\pm$ 0.39 & 54.72 $\pm$ 0.33 & 99.44 $\pm$ 0.08 & 78.68\\
W/O Hopfield & 81.03 $\pm$ 0.45 & 54.96 $\pm$ 0.28 & 99.44 $\pm$ 0.03 & 78.48 \\
W/O Memory & 79.53 $\pm$ 0.36 & 53.19 $\pm$ 0.37 & 99.47 $\pm$ 0.07 & 77.40\\
W/O Bottleneck & 75.40 $\pm$ 0.48 & 46.53 $\pm$ 0.41 & 93.33 $\pm$ 0.15 & 73.75\\
W/O SA & 72.72 $\pm$ 0.57 & 47.75 $\pm$ 0.31 & 99.46 $\pm$ 0.04 & 73.31\\
\bottomrule
\end{tabular}
\caption{The ablation study demonstrated that leveraging all the components resulted in the best performance.}
\label{tab:abla}
\end{table*}

\begin{table}[t]
  \centering
  \small
  \begin{tabular}{lc}
    \toprule
    Models & \#FLOPs \\
    \hline
    AiT & 9.64$\times 10^8$ \\
    Hopfield network component & 8.02$\times 10^6$ \\
    W/O Hopfield (replaced with cross-attention) & 1.19$\times 10^9$ \\
    \bottomrule
  \end{tabular}
    \caption{Hopfield network FLOPs.}
     \label{tab:flops_comparison}
\end{table}

\begin{table}
  \centering
  \small
  \begin{tabular}{lc}
    \toprule
    Initialization & Accuracy (\%) \\
    \hline
    Gaussian distribution & \textbf{83.34 $\pm$ 0.44}\\
    Uniform distribution \cite{turing} & 81.92 $\pm$ 0.17\\
    Identity distribution \cite{goyal} & 78.56 $\pm$ 0.29\\
    \bottomrule
  \end{tabular}
    \caption{Memory initialization methods.}
  \label{tab:init}
\end{table}

A comprehensive ablation study is aimed to acquire insights into the functionalities of different components in AiT. In particular, we conducted the following ablations:
\begin{itemize}[left=0pt]
    \setlength\itemsep{0em}
    \item Reset Memory: The priors in the explicit memory are initialized every epoch.
    \item W/O Hopfield: The Hopfield network is removed and replaced with cross-attention that shares the same architecture as the multi-head cross-attention in Figure \ref{fig:sche}.a. The cross-attention takes the input $\Xi^t$ as the query and the upscaled priors $f_{\text{LT}}(\gamma^{t+1})$ as the key and value to compute the output $\hat{\Xi}^{t} = \text{MHA}(\Xi^t,f_{\text{LT}}(\gamma^{t+1}))$. The rationale behind this ablation is grounded in studies that relied on iterative cross-attentions, such as Set Transformer. 
    \item W/O Memory: The global workspace layer is removed.
    \item W/O Attention Balance Loss: The bottleneck is implemented without the attention balance loss.
    \item W/O Bottleneck: The top-$k$ bottleneck attention is replaced with dense pair-wise attention.
    \item W/O SA: The self-attention in Figure \ref{fig:sche}.b is removed.
\end{itemize}
Table \ref{tab:abla} demonstrates that the bottleneck plays a significant role in enhancing performance, where its absence led to a sizable decrease in accuracy. The \lq W/O Attention Balance Loss\rq ablation presents the effectiveness of the proposed bottleneck attention balance loss. Ablating the other components, such as the Hopfield networks and the explicit memory, while not as impactful, still resulted in degraded accuracy. Consequently, the complete model with all the components achieved the best performance. Additionally, for the \lq W/O Hopfield\rq ablation, Table \ref{tab:flops_comparison} demonstrates the training time \#FLOPs of the Hopfield network compared to the cross-attention component. The Hopfield network is a lightweight component utilizing less than 0.84\% FLOPs of the proposed AiT model.

\subsection{Explicit memory initialization}

In the explicit memory, we initialized priors with values drawn from a specific distribution, including the Gaussian distribution, the uniform distribution \cite{turing}, and the identity distribution \cite{goyal}. In particular, the Gaussian distribution generates random values with a mean of zero and a variance of one. The uniform distribution utilizes an upper bound of $\frac{1}{\sqrt{M+D}}$, where $M$ denotes the number of priors, and $D$ is the dimension of a prior. The identity distribution assigns ones on the diagonal and zeros elsewhere. Table \ref{tab:init} indicates that the Gaussian distribution resulted in the best performance, which potentially prevents specific priors from dominating the learning process, for a more balanced memory update in the early training stages.

\subsection{Parameter efficiency}

Benefits of incorporating the Global Workspace Layer include enhanced model performance and parameter efficiency. We conducted extensive experiments on AiT's parameter efficiency by measuring the model size and test accuracy in CIFAR-10. In particular, we compared a single AiT block to a Coordination block that consists of two multi-head cross-attention components. We explored various configurations by replacing the cross-attention components in the Coordination with the proposed low-rank memory and Hopfield network in AiT. 

\begin{figure}
    \centering
    \includegraphics[width=0.8\linewidth]{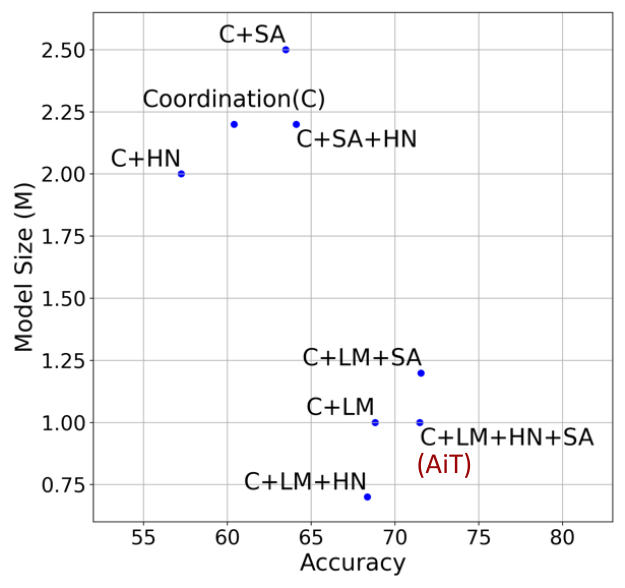}
    \caption{Model size vs. test accuracy for various model configurations. Consolidating all the components in the Coordination block resulted in the best performance of 71.49\% maintaining a compact model size of 1.0M.}
    \label{fig:coor}
\end{figure}

In Figure \ref{fig:coor}, \lq C\rq$\,$denotes the default Coordination block, \lq LM\rq$\,$denotes replacing the first cross-attention in the Coordination block with the low-rank explicit memory, \lq HN\rq$\,$denotes replacing the second cross-attention with the Hopfield network, and \lq SA\rq$\,$denotes the self-attention component. Compared to the Coordination (C) which achieved an accuracy of 60.41\% with a model size of 2.2M, integrating the low-rank memory (C+LM) significantly improved accuracy while reducing the model size to 1M. Moreover, the Hopfield network (HN) achieved competitive performance in C+SA+HN and C+LM+HN+SA configurations for enhanced parameter efficiency. In addition, HN appeared to be effective only when the LM or SA component was employed. This is attributed to the retrieval within the associative memory of the Hopfield network relying on learning a diverse set of priors, which are learned based on the low-rank memory and the self-attention components. Removing these components decreases the model's ability to learn meaningful priors. Consequently, consolidating all the three components in the Coordination block (C+LM+HN+SA), i.e., an AiT block, resulted in the best performance of 71.49\% maintaining a compact size of 1.0M. AiT is a sparse representation learner, surpassing the Coordination method in terms of parameter efficiency.

\subsection{Prior specialization}

The bottleneck attention balance loss encourages competition facilitating the selection of diverse tokens through the bottleneck. We aim to investigate whether the bottleneck attention could contribute to specialized priors that guide attention to various localized features for long-range association. To quantitatively measure the diversity of the selected tokens, we computed the ratio of distinct instances in the selected tokens. Notably, for each prior in the explicit memory, we visualized its attention scores computed by Equation \ref{eq:attention} over the different input token positions. Figure \ref{fig:spar} presents the corresponding image patches for the selected tokens, with an apparent increase in the patches' diversity as training progresses. Additionally, Figure \ref{fig:spec} highlights these attended patches in the image samples where the learned priors demonstrated emergent spatial specialization, dividing labor by focusing on various aspects of the images.

\subsection{Relational reasoning tasks}
\label{sec:rel}

The specialization in the spatial attention allocation of priors results in AiT's enhanced performance in relational reasoning tasks. The aim is to answer questions about the properties and relations of various objects in a given image. 
We evaluated AiT's performance in the Sort-of-CLEVR dataset \cite{sort} for the relational reasoning tasks, which comprises both non-relational and relational tasks. In the non-relational tasks, the question involves the attributes of a specific object, whereas in the relational tasks, the question focuses on the relations between different objects. 

\begin{table}
\centering
\small
\begin{tabular}{l|cc}
\toprule
Methods & Relational (\%) & Non-relational (\%)\\
\hline
\multicolumn{3}{l}{Transformer based models} \\
\hline 
AiT-Base & \textbf{80.03} & \textbf{99.98}\\
AiT-Medium & 78.14 & 99.75\\
AiT-Small & 76.82 & 99.85\\
Coordination \cite{goyal} & 73.43 & 96.31\\ %
ViT-Base & 63.35 & 99.73 \\
ViT-Medium & 54.71 & 99.70\\
ViT-Small & 51.75 & 98.80\\
Set Transformer \cite{lee19} & 47.63 & 57.65\\ %
\hline
\multicolumn{3}{l}{Non-Transformer based models} \\
\hline
CNN+LSTM \cite{sort} & 60.08 & 99.47\\ 
Dense-Base & 46.93 & 57.71\\  
Dense-Small & 47.28 & 57.68\\  
\bottomrule
\end{tabular}
\caption{Relational reasoning tasks.}
\label{tab:sort}
\end{table}

To tackle the Sort-of-CLEVR task, we embedded a question using a learnable linear projection, with layer normalization employed before and after the embedding. Then, the question embedding was concatenated to the image patch embeddings as the input to AiT. We compared to both Transformer-based models including Set Transformer and Coordination, and other non-Transformer based models including CNN+LSTM \cite{sort} and Dense networks. In CNN+LSTM, images were processed with a CNN to produce a set of objects, and questions were processed with an LSTM to generate a question embedding. Then, an MLP combined the image and question embeddings to predict the answer of the question.
Dense-Small and Dense-Base are two fully-connected models derived from AiT-Small and AiT-Base. Table \ref{tab:sort} demonstrates AiT's superior performance in the relational and non-relational tasks compared to the Coordination and Set Transformer, achieving a competitive accuracy of 80.03\% in the relational tasks. In addition, AiT outperformed the non-Transformer CNN+LSTM model, which relied on a stronger assumption of the input data, i.e., employing the CNN and LSTM for the images and questions, respectively. By contrast, AiT utilized the concatenated tokens of images and questions for the same operations in the proposed Global Workspace Layer.

\section{Conclusions} 

We proposed the Associative Transformer (AiT) based on the bottleneck attention in the explicit memory and the token reconstruction within associative memory to enhance long-range associations of sparsely attended tokens in sparse Transformers. The extensive experiment results demonstrated AiT's enhanced performance and parameter efficiency in the various classification and relational reasoning tasks, surpassing the Vision Transformers and conventional sparse Transformers. We are excited about inducing associative memory for localized contextual learning in sparse Transformers through the set of diverse priors and the continuous Hopfield networks for various tasks.

\paragraph{Limitations}
Scaling AiT to larger dataset training without fine-tuning the Global Workspace Layer's hyperparameters such as the Hopfield inverse temperature remains a challenge. Implementing a mechanism to adjust adaptively could enhance model performance as it scales. We aim to investigate the countermeasure for more complex tasks, including the CLEVER \cite{clevr} and bAbI \cite{weston2015towards} datasets in the future. AiT enables more accessible and cost-effective Transformer model training, showing promise for applications in other domains, such as audio and video. However, potential risks include susceptibility to targeted adversarial attacks, such as Trojan samples exploiting the sparse bottleneck attention mechanism. Exploring potential countermeasures is crucial for its safe deployment in real-world applications. 

\section*{Acknowledgment}
This work was supported in part by Microsoft Research Asia D-CORE Collaborative Research Program.

\bibliography{references}

\begin{thebibliography}{10}

\bibitem{attention}
Ashish Vaswani, Noam Shazeer, Niki Parmar, and et~al.
\newblock Attention is all you need.
\newblock {\em NeurIPS}, pages 5998--6008, 2017.

\bibitem{vit}
Alexey Dosovitskiy, Lucas Beyer, Alexander Kolesnikov, Dirk Weissenborn, Xiaohua Zhai, Thomas Unterthiner, Mostafa Dehghani, Matthias Minderer, Georg Heigold, Sylvain Gelly, et~al.
\newblock An image is worth 16x16 words: Transformers for image recognition at scale.
\newblock {\em arXiv preprint arXiv:2010.11929}, 2020.

\bibitem{brook91}
Rodney~A. Brooks.
\newblock Intelligence without representation.
\newblock {\em Artif. Intell.}, 47(1-3):139--159, 1991.

\bibitem{binding}
Klaus Greff, Sjoerd van Steenkiste, and J{\"{u}}rgen Schmidhuber.
\newblock On the binding problem in artificial neural networks.
\newblock {\em arXiv preprint arXiv:2012.05208}, 2020.

\bibitem{Baars1988}
Bernard~J. Baars.
\newblock {\em A Cognitive Theory of Consciousness}.
\newblock Cambridge University Press, 1988.

\bibitem{dehaene98}
Stanislas Dehaene, Michel Kerszberg, and Jean Changeux.
\newblock A neuronal model of a global workspace in effortful cognitive tasks.
\newblock {\em Proceedings of the National Academy of Sciences}, 95(24):14529--14534, 1998.

\bibitem{vanr20}
Rufin VanRullen and Ryota Kanai.
\newblock Deep learning and the global workspace theory.
\newblock {\em arXiv preprint arXiv:2012.10390}, 2020.

\bibitem{juliani2022}
Arthur Juliani, Kai Arulkumaran, Shuntaro Sasai, and Ryota Kanai.
\newblock On the link between conscious function and general intelligence in humans and machines.
\newblock {\em Transactions on Machine Learning Research}, 2022.

\bibitem{cog2}
Edward Awh, Edward~K Vogel, and S-H Oh.
\newblock Interactions between attention and working memory.
\newblock {\em Neuroscience}, 139(1):201--208, 2006.

\bibitem{cog1}
Adam Gazzaley and Anna~C Nobre.
\newblock Top-down modulation: bridging selective attention and working memory.
\newblock {\em Trends in cognitive sciences}, 16(2):129--135, 2012.

\bibitem{nfl1}
Jonathan Baxter.
\newblock A model of inductive bias learning.
\newblock {\em Journal of artificial intelligence research}, 12:149--198, 2000.

\bibitem{nfl2}
Anirudh Goyal and Yoshua Bengio.
\newblock Inductive biases for deep learning of higher-level cognition.
\newblock {\em Proceedings of the Royal Society A}, 478(2266):20210068, 2022.

\bibitem{goyal}
Anirudh Goyal, Aniket~Rajiv Didolkar, Alex Lamb, and et~al.
\newblock Coordination among neural modules through a shared global workspace.
\newblock In {\em ICLR}, 2022.

\bibitem{sort}
Adam Santoro, David Raposo, David G.~T. Barrett, and et~al.
\newblock A simple neural network module for relational reasoning.
\newblock In {\em {NIPS}}, pages 4967--4976, 2017.

\bibitem{shanahan2020explicitly}
Murray Shanahan, Kyriacos Nikiforou, Antonia Creswell, Christos Kaplanis, David Barrett, and Marta Garnelo.
\newblock An explicitly relational neural network architecture.
\newblock In {\em International Conference on Machine Learning}, pages 8593--8603. PMLR, 2020.

\bibitem{mittal2021compositional}
Sarthak Mittal, Sharath~Chandra Raparthy, Irina Rish, Yoshua Bengio, and Guillaume Lajoie.
\newblock Compositional attention: Disentangling search and retrieval.
\newblock {\em arXiv preprint arXiv:2110.09419}, 2021.

\bibitem{spies2022sparse}
Alex~F Spies, Alessandra Russo, and Murray Shanahan.
\newblock Sparse relational reasoning with object-centric representations.
\newblock {\em arXiv preprint arXiv:2207.07512}, 2022.

\bibitem{kerg2022neural}
Giancarlo Kerg, Sarthak Mittal, David Rolnick, Yoshua Bengio, Blake Richards, and Guillaume Lajoie.
\newblock On neural architecture inductive biases for relational tasks.
\newblock {\em arXiv preprint arXiv:2206.05056}, 2022.

\bibitem{jiang2023object}
Jindong Jiang, Fei Deng, Gautam Singh, and Sungjin Ahn.
\newblock Object-centric slot diffusion.
\newblock {\em arXiv preprint arXiv:2303.10834}, 2023.

\bibitem{mondal2024slot}
Shanka~Subhra Mondal, Jonathan~D Cohen, and Taylor~W Webb.
\newblock Slot abstractors: Toward scalable abstract visual reasoning.
\newblock {\em arXiv preprint arXiv:2403.03458}, 2024.

\bibitem{johnhf}
John~J. Hopfield.
\newblock Hopfield network.
\newblock {\em Scholarpedia}, 2(5):1977, 2007.

\bibitem{hfl}
Hubert Ramsauer, Bernhard Sch{\"{a}}fl, Johannes Lehner, and et~al.
\newblock Hopfield networks is all you need.
\newblock In {\em ICLR}, 2021.

\bibitem{lee19}
Juho Lee, Yoonho Lee, Jungtaek Kim, and et~al.
\newblock Set transformer: {A} framework for attention-based permutation-invariant neural networks.
\newblock In {\em {ICML}}, pages 3744--3753, 2019.

\bibitem{gmat}
Ankit Gupta and Jonathan Berant.
\newblock {GMAT:} global memory augmentation for transformers.
\newblock {\em arXiv preprint arXiv:2006.03274}, 2020.

\bibitem{luna}
Xuezhe Ma, Xiang Kong, Sinong Wang, and et~al.
\newblock Luna: Linear unified nested attention.
\newblock In {\em NeurIPS}, pages 2441--2453, 2021.

\bibitem{jaegle21}
Andrew Jaegle, Felix Gimeno, Andy Brock, and et~al.
\newblock Perceiver: General perception with iterative attention.
\newblock In {\em {ICML}}, volume 139 of {\em Proceedings of Machine Learning Research}, pages 4651--4664. {PMLR}, 2021.

\bibitem{jaegle22}
Andrew Jaegle, Sebastian Borgeaud, Jean{-}Baptiste Alayrac, and et~al.
\newblock Perceiver {IO:} {A} general architecture for structured inputs {\&} outputs.
\newblock In {\em {ICLR}}, 2022.

\bibitem{qiu20}
Jiezhong Qiu, Hao Ma, Omer Levy, and et~al.
\newblock Blockwise self-attention for long document understanding.
\newblock In {\em {EMNLP} (Findings)}, pages 2555--2565, 2020.

\bibitem{modular3}
Simiao Zuo, Xiaodong Liu, Jian Jiao, and et~al.
\newblock Taming sparsely activated transformer with stochastic experts.
\newblock In {\em {ICLR}}, 2022.

\bibitem{modular4}
James~Urquhart Allingham, Florian Wenzel, Zelda~E. Mariet, and et~al.
\newblock Sparse moes meet efficient ensembles.
\newblock {\em Transactions on Machine Learning Research}, 2022.

\bibitem{turing}
Alex Graves, Greg Wayne, and Ivo Danihelka.
\newblock Neural turing machines.
\newblock {\em arXiv preprint arXiv:1410.5401}, 2014.

\bibitem{modernhf}
Dmitry Krotov and John~J. Hopfield.
\newblock Dense associative memory for pattern recognition.
\newblock In {\em {NIPS}}, pages 1172--1180, 2016.

\bibitem{chan18}
{\c{C}}aglar G{\"{u}}l{\c{c}}ehre, Sarath Chandar, Kyunghyun Cho, and Yoshua Bengio.
\newblock Dynamic neural turing machine with continuous and discrete addressing schemes.
\newblock {\em Neural Comput.}, 30(4), 2018.

\bibitem{demi}
Mete Demircigil, Judith Heusel, Matthias Löwe, and et~al.
\newblock On a model of associative memory with huge storage capacity.
\newblock {\em Journal of Statistical Physics}, 168(2):288--299, 2017.

\bibitem{hoover23}
Benjamin Hoover, Yuchen Liang, Bao Pham, and et~al.
\newblock Energy transformer.
\newblock {\em arXiv preprint arXiv:2302.07253}, 2023.

\bibitem{cifar100}
Alex Krizhevsky and Geoffrey Hinton.
\newblock Learning multiple layers of features from tiny images.
\newblock Technical report, 2009.

\bibitem{pet}
Omkar~M. Parkhi, Andrea Vedaldi, Andrew Zisserman, and C.~V. Jawahar.
\newblock Cats and dogs.
\newblock In {\em CVPR}, pages 3498--3505, 2012.

\bibitem{deng2009imagenet}
Jia Deng, Wei Dong, Richard Socher, Li-Jia Li, Kai Li, and Li~Fei-Fei.
\newblock Imagenet: A large-scale hierarchical image database.
\newblock In {\em 2009 IEEE conference on computer vision and pattern recognition}, pages 248--255. Ieee, 2009.

\bibitem{mittal20}
Sarthak Mittal, Alex Lamb, Anirudh Goyal, Vikram Voleti, Murray Shanahan, Guillaume Lajoie, Michael Mozer, and Yoshua Bengio.
\newblock Learning to combine top-down and bottom-up signals in recurrent neural networks with attention over modules.
\newblock In {\em International Conference on Machine Learning}, pages 6972--6986. PMLR, 2020.

\bibitem{clevr}
Justin Johnson, Bharath Hariharan, Laurens van~der Maaten, and et~al.
\newblock {CLEVR:} {A} diagnostic dataset for compositional language and elementary visual reasoning.
\newblock In {\em CVPR}, pages 1988--1997, 2017.

\bibitem{weston2015towards}
Jason Weston, Antoine Bordes, Sumit Chopra, Alexander~M Rush, Bart Van~Merri{\"e}nboer, Armand Joulin, and Tomas Mikolov.
\newblock Towards ai-complete question answering: A set of prerequisite toy tasks.
\newblock {\em arXiv preprint arXiv:1502.05698}, 2015.

\end{thebibliography}
\bibliographystyle{unsrt}

\newpage
\appendix
\onecolumn
\section{Datasets}
\label{appe:data}

\begin{figure}[!h]
    \centering
    \includegraphics[width=0.6\linewidth]{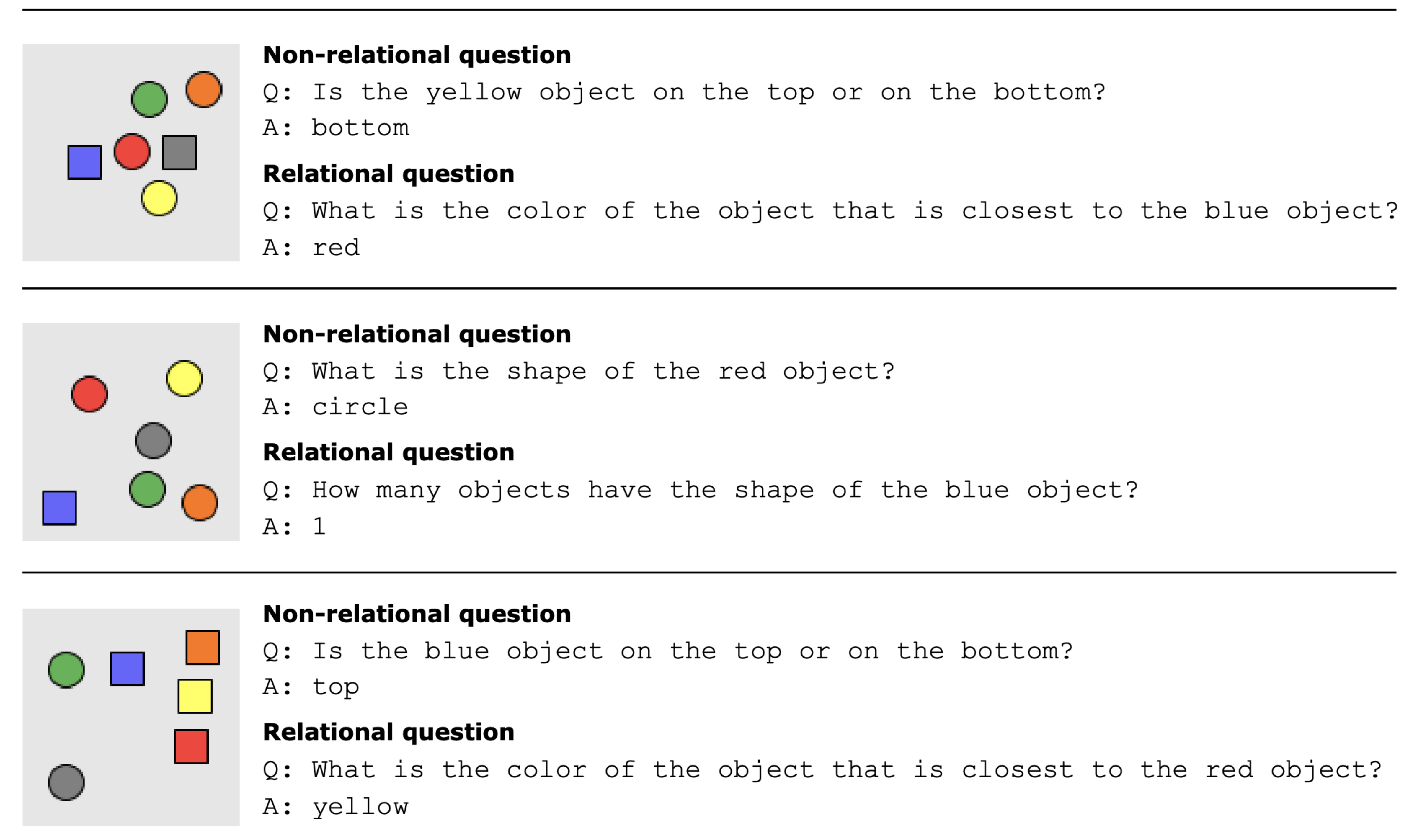}
    \caption{Examples from the Sort-of-CLEVR dataset \cite{sort}.}
    \label{fig:appesort}
\end{figure}

In this section, we describe the datasets used in this work. (1) CIFAR-10 is an image collection of 10 objects, covering 50k training samples and 10k test samples, labeled as airplane, automobile, and so on. The size of images is $32\times32\times3$. (2) CIFAR-100 \cite{cifar100} contains 100 object classes with 500 training images and 100 testing images per class. For both the CIFAR-10 and CIFAR-100 datasets, we performed random cropping with size $32\times32\times3$ and a padding size of 4. (3) Triangle dataset \cite{goyal} includes 50k training images and 10k test images with size $64\times 64$, each of which contains 3 randomly placed clusters of points. The task is to predict whether the three clusters form an equilateral triangle or not. (4) Oxford-IIIT Pet dataset \cite{pet} comprises 37 categories featuring diverse breeds of cats and dogs, with 200 images allocated for each class. We utilized random resized cropping with size $256\times256\times3$ and resized all images to size $224\times224\times3$. Additionally, we applied random horizontal flip and normalization to the CIFAR-10, CIFAR-100, and Oxford-IIIT Pet datasets. (5) Sort-of-CLEVR dataset \cite{sort} is a simplified version of the CLEVR dataset \cite{clevr}. It includes 10k images with size $75\times75\times3$ and 20 different questions (10 relational and 10 non-relational questions) for each image. In each image, objects with randomly chosen shapes (square or circle) and randomly chosen colors (red, green, blue, orange, gray, yellow) are placed (Figure \ref{fig:appesort}).

\section{Comparison of work related to Global Workspace Theory}
\label{appe:rela}

We discuss and summarize the existing sparse attention methods in relation to the properties of Global Workspace Theory (Table \ref{tab:related work}). First, we examine whether an architecture involves operations of information writing and reading through a shared workspace. Secondly, we assess whether the latent representations (priors) in workspace memory are subsequently processed by self-attention. Thirdly, we inspect whether the latent representations have a lower rank compared to the input representations. Fourthly, we analyze whether information retrieval from the workspace is driven by a bottom-up or a top-down signal. Lastly, we investigate whether the model incorporates a bottleneck with a limited capacity to regulate the information flow passing through the workspace.

\begin{table}[h]
\centering
\small
\renewcommand{\arraystretch}{1.05}
\resizebox{\textwidth}{!}{%
\begin{tabular}{l|cc|c|c|c|c}
\toprule
\multirow{2}{*}{Method} & \multicolumn{2}{c|}{Operations} & \multirow{2}{*}{Self-Attention} & \multirow{2}{*}{Low-Rank Memory} & \multirow{2}{*}{Top-Down/Bottom-Up} & \multirow{2}{*}{Bottleneck} \\
\cline{2-3}
& Writing & Reading & & & & \\
\hline
Vision Transformer \cite{vit} & $\times$ & $\times$ & $\checkmark$ & $\times$ & BU & $\times$ \\
BlockBERT \cite{qiu20} & $\times$ & $\times$ & $\checkmark$ & $\times$ & BU & $\checkmark$\\
BRIMs\cite{mittal20} & $\times$ & $\times$ & $\times$ & $\times$ & TD & $\checkmark$\\
Modern Hopfield \cite{hfl} & $\times$ & $\checkmark$ & $\times$ & $\checkmark$ & BU & $\times$ \\
Perceiver \cite{jaegle21} & $\checkmark$ & $\times$ & $\checkmark$ & $\checkmark$ & BU & $\times$ \\
Coordination \cite{goyal} & $\checkmark$ & $\checkmark$ & $\times$ & $\times$ & BU & $\checkmark$ \\
Perceiver IO \cite{jaegle22}& $\checkmark$ & $\checkmark$ & $\checkmark$ & $\checkmark$ & TD & $\times$ \\
Set Transformer \cite{lee19} & $\checkmark$ & $\checkmark$ & $\times$ & $\times$ & BU & $\times$ \\
Luna \cite{luna} & $\checkmark$ & $\checkmark$ & $\times$ & $\times$ & BU & $\times$\\
GMAT \cite{gmat} & $\checkmark$ & $\checkmark$ & $\checkmark$ & $\times$ & BU & $\checkmark$\\
Associative Transformer (Ours) & $\checkmark$ & $\checkmark$ & $\checkmark$ & $\checkmark$ & BU & $\checkmark$\\
\bottomrule
\end{tabular}%
}
\caption{Comparison of attention architectures based on properties of the Global Workspace Theory.}
\label{tab:related work}
\end{table}

\section{Experimental settings and hyperparameters}
\label{appe:hype}

Table \ref{tab:hyperparameters} presents the hyperparameters used for the different tasks in this study. Unless otherwise noted, we employed the author-recommended settings and hyperparameters for the re-implementation of baseline models. The small variants of ViT and AiT have 2 attention layers, and the base variants of them have 12 attention layers instead. We used the same dimension of the hidden layer and the MLP layer for ViT and AiT. By default, we employed 8 attention heads and 32 memory slots for the bottleneck attention. To obtain the bottleneck size, we considered two main factors of the batch size and the patch size. For the CIFAR, Pet, and ImageNet100 datasets, we used a bottleneck size of 512, which selected from a pool of 32.8k/25.1k tokens. For the Triangle dataset, we used a bottleneck size of 64 from a pool of 2.0k tokens. For the relational reasoning tasks, we used a bottleneck size of 256, which selected from a pool of 14.4k tokens. Based on the bottleneck size and the patch pool size, we used 128 memory slots for the Pet and ImageNet100 datasets and 32 memory slots for the other datasets. Moreover, we trained the models on the Pet dataset for 300 epochs and on the other datasets for 100 epochs. In relational reasoning tasks, we trained all models for 100 epochs with a batch size of 64.

\begin{table}[!h]
\centering
\small
\renewcommand{\arraystretch}{1.2}
\begin{tabular}{p{6cm}p{4cm}}
\toprule
\textbf{Parameter} & \textbf{Value} \\
\hline
\multicolumn{2}{l}{\textbf{Common parameters}} \\
\hline
Optimizer & AdamW \\
Weight decay & 0.01 \\
Learning rate & $1 \times 10^{-4}$ \\
Number of self-attention heads & 12 \\
Number of attention layers & 2 (Small)/ 12 (Base)\\
Size of hidden layer & 768 \\
Size of MLP & 3072 \\
Size of memory slot & 32 \\
Number of bottleneck attention heads & 8 \\
Beta & 1.0 \\
Epochs & 100 (300 for Oxford Pet)\\
\hline
\multicolumn{2}{l}{\textbf{CIFAR}} \\
\hline
Patch size & 4 \\
Batch size & 512 \\
Number of memory slots & 32 \\
Bottleneck size & 512 \\
\hline
\multicolumn{2}{l}{\textbf{Triangle}} \\
\hline
Patch size & 32 \\
Batch size & 512 \\
Number of memory slots & 32 \\
Bottleneck size & 64 \\
\hline
\multicolumn{2}{l}{\textbf{Oxford Pet and ImageNet100}} \\
\hline
Patch size & 16 \\
Batch size & 128 \\
Number of memory slots & 128 \\
Bottleneck size & 512 \\
\hline
\multicolumn{2}{l}{\textbf{Relational reasoning}} \\
\hline
Patch size & 5 \\
Batch size & 64 \\
Number of memory slots & 32 \\
Bottleneck size & 256 \\
\bottomrule
\end{tabular}
\caption{Hyperparameters.}
\label{tab:hyperparameters}
\end{table}

\section{Algorithm}
\label{appe:algo}

\begin{algorithm}[h]
\small
\caption{Global Workspace Layer}
\begin{algorithmic}[1] 
\STATE \textbf{Main program}
\STATE \textbf{Input:} tokens from the previous layers: $V\in\mathbb{R}^{B\times N\times E}$; learnable low-rank priors: $\gamma \in \mathbb{R}^{M\times D}$
\STATE Squash layer concatenates tokens in the batch:   $\Xi\in\mathbb{R}^{BN\times E} \leftarrow V\in\mathbb{R}^{B\times N\times E}$
\STATE Project $\Xi$ to a latent space with a dimension $D\ll E$:   $\Xi W^K_{i} \in\mathbb{R}^{BN\times D}$
\STATE Obtain the attention scores over the projected tokens using priors $\gamma$:   $\text{A}_i(\gamma, \Xi)=\text{softmax}(\frac{\gamma(\Xi W^K_{i})^T}{\sqrt{D}})$ 
\STATE Tokens compete to write in memory via a bottleneck with capacity $k$:  $h_i = \text{top-}k(\text{A}_i)\Xi W^V$ (*Bottleneck Attention Balance Loss is employed for a more diverse token selection)
\STATE Update priors $\gamma^t$ with Exponentially Weighted Moving Average:  $\hat{\gamma}^{t+1} = (1 - \alpha) \cdot \gamma^{t} + \alpha \cdot \text{LN}(\text{Concat}(h_1,\dots,h_S)W^O)$
\STATE Layer normalization:  $\gamma^{t+1} = \frac{{\hat{\gamma}^{t+1}}}{{\sqrt{\sum_{{j=1}}^{M} (\hat{\gamma}^{t+1}_j)^2}}}$
\STATE Project $\gamma^{t+1} \in \mathbb{R}^{M\times D}$ into a dimension of $E$ as attractors within associative memory: $f_{\text{LT}}(\gamma^{t+1}) \in \mathbb{R}^{M\times E}$
\STATE Reconstruct $\Xi$ using attractors $f_{\text{LT}}(\gamma^{t+1})$ with a continuous Hopfield network in one inner step of energy reduction:  
 $\hat{\Xi}^t = \arg\min_{\Xi^t} (-\text{lse}(\beta,f_{\text{LT}}(\gamma^{t+1}) \Xi^t)+\frac{1}{2}\Xi^t{\Xi^t}^T+
    \beta^{-1}\text{log}M+\frac{1}{2}(\max_{i} |f_{\text{LT}}(\gamma^{t+1}_i)|)^2)$
\STATE Reshape the tokens into batches as the output of the Global Workspace layer:  $\hat{V}^t \in \mathbb{R}^{B \times N \times E} \leftarrow \hat{\Xi}^t \in \mathbb{R}^{(B \times N) \times E}$
\STATE \textbf{Bottleneck Attention Balance Loss}
\STATE Cumulative attention loss: $\ell_{\text{importance}_{i,o}}  = \sum_{j=1}^M \text{A}_{i,j,o}$
\STATE Selected instance loss: $\ell_{\text{loads}_{i,o}}  = \sum_{j=1}^M (\text{A}_{i,j,o} > 0)$
\STATE For each attention head $i$: $\ell_{\text{bottleneck}_{i}}  = \frac{{\text{Var}(\{\ell_{\text{importance}_{i,o}}}\}_{o=1}^{B \times N})}{{(\frac{1}{B \times N}\sum_{o=1}^{B \times N}\ell_{\text{importance}_{i,o}}})^2 + \epsilon} + \frac{{\text{Var}(\{\ell_{\text{loads}_{i,o}}}\}_{o=1}^{B \times N})}{{(\frac{1}{B \times N}\sum_{o=1}^{B \times N}\ell_{\text{loads}_{i,o}}})^2 + \epsilon}$
\STATE Sum the losses over all heads: $\sum_{i=1}^S \ell_{\text{bottleneck}_i}$
\end{algorithmic}
\end{algorithm}

\section{Analysis of attention head operating modes}

\begin{figure}[!h]
    \centering
    \includegraphics[width=0.75\linewidth]{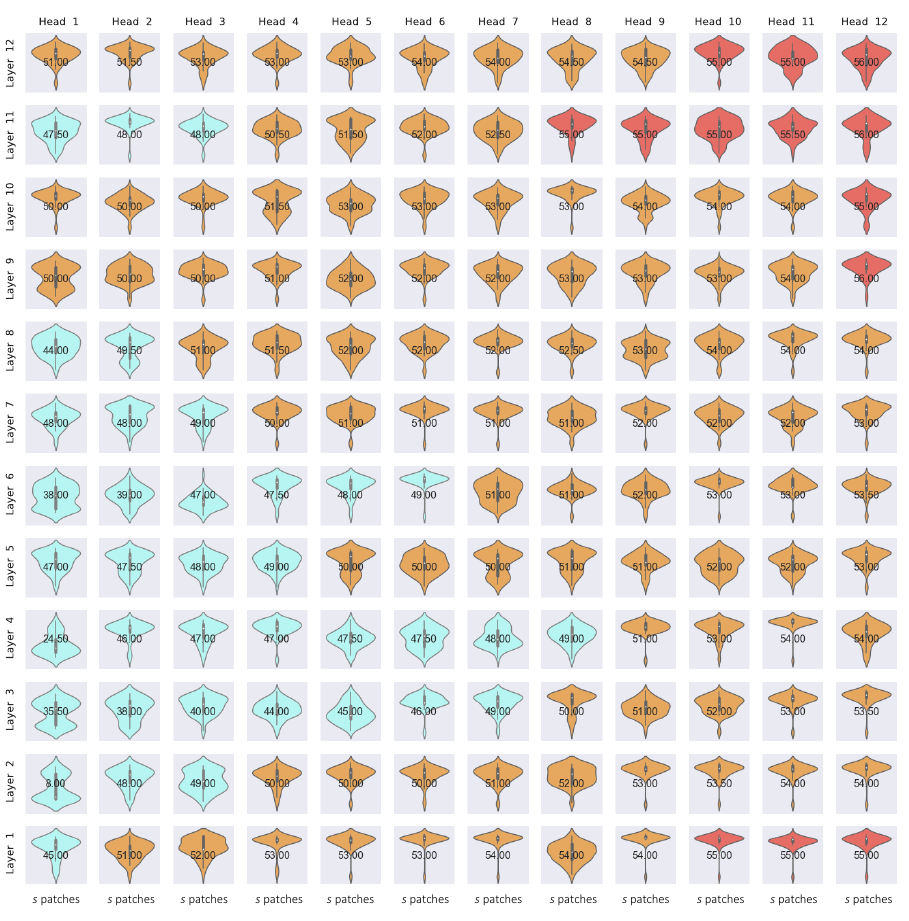}
    \caption{Analysis of operating modes of attention heads in the ViT-Base model. We recognize three different groups of attention heads based on their sparsity scores. Group (I) in light blue: High sparsity heads abundant in the middle layers 3-6. The vast majority of these heads only used 50\% or fewer interactions. Group (II) in orange: Middle sparsity heads predominant in layers 2 and 7-10. Less than 80\% of the interactions were activated. Group (III) in red: Low sparsity heads observed in high layers 11-12 and the first layer, where the most tokens were attended to. The global workspace layer will provide the inductive bias to attend to the essential tokens more effectively.}
    \label{fig:head}
\end{figure}

We assume that the competition within the pair-wise attention is important for the model to learn meaningful representations. If such competition exists, a trained model will naturally result in sparser interactions in attention heads. Therefore, we first performed an analysis of the operating modes of different attention heads in a pretrained ViT model by measuring the number of tokens each head is attending to. We trained a ViT-Base variant model for 100 epochs from scratch for the CIFAR-10 task. Then, for each attention head, we obtained a violin plot to represent the distribution of attention sparsity for different tokens (Figure \ref{fig:head}). The attention sparsity for a specific patch's interactions with other tokens is computed as follows $\arg\min_{s} \sum_{j=1}^{s} A^{i,j} \geq 0.9$, where $A^{i,j}$ is the attention score allocated to the $j$th patch by the $i$th patch. The attention sparsity score is measured by the minimal number of required tokens whose attention scores add up to 0.90. For instance, there are 65 tokens for the CIFAR-10 task with patch size 4, thus there are 65 interactions for each patch to all the tokens including itself. An attention head has a higher sparsity if the median $\bar{s}$ of the required tokens is smaller, which is the number in the center of each panel. The heads in each layer are sorted according to $\bar{s}$. Note that training the model for a longer duration can result in even better convergence and higher attention sparsity. We also refer to a concurrent investigation on the Bidirectional Encoder Representations from Transformers (BERT) for results on the natural language processing (NLP) tasks \cite{hfl}. 

Compared to NLP tasks, much less sparsity was found in ViT models. This can be attributed to the long-range dependencies of the attention mechanism were more frequently employed in handling patches of images, compared to handling tokens in text sequences. In ViT models, an input image is divided into non-overlapping patches which are then treated as separate "tokens" and fed into the transformer model. By contrast, the sequential nature of the NLP text data allows the model to capture long-range dependencies among long-range tokens more easily. Furthermore, depending on the observation of sparsity, the bottleneck size proposed in this work can vary across layers in practice, and different datasets with varying complexities require different capacities as well. When a model has fewer layers, a constant bottleneck size can perform competitively with a well-tuned one. However, for deeper models, further investigation is needed to determine the optimal capacity size. For example, middle layers might benefit more from a smaller bottleneck compared to other layers.

\section{Ablation study during test time}
\label{appe:test}
During the test time, we included a new \lq W/O Attention\rq$\,$ ablation, where the multi-head cross-attention is disabled, and only the Hopfield networks are leveraged for the retrieval from the explicit memory. We evaluated ViT performance in both image classification (with AiT-Medium) and relational reasoning tasks (with AiT-Small). Table \ref{tab:test1} showed that the multi-head cross-attention with the bottleneck is important for improving AiT's performance during the test time.

\begin{table}[!h]
\centering
\small
\begin{tabular}{l|ccccc}
\toprule
Methods & CIFAR10 (\%)& CIFAR100 (\%)& Triangle (\%)& Oxford Pet (\%)& Relational tasks (\%)\\
\hline
AiT & \textbf{84.59} & \textbf{60.58} & \textbf{99.57} & \textbf{30.05} & \textbf{76.82}\\
W/O Attention & 84.50 & 60.56 & 99.56 & 28.68 & 75.17 \\
\bottomrule
\end{tabular}
\caption{Performance comparison with the W/O Attention ablation.}
\label{tab:test1}
\end{table}

\section{Efficacy of the bottleneck attention balance loss}
\label{appe:bott}

\begin{figure}[!h]
    \centering
    \includegraphics[width=0.6\linewidth]{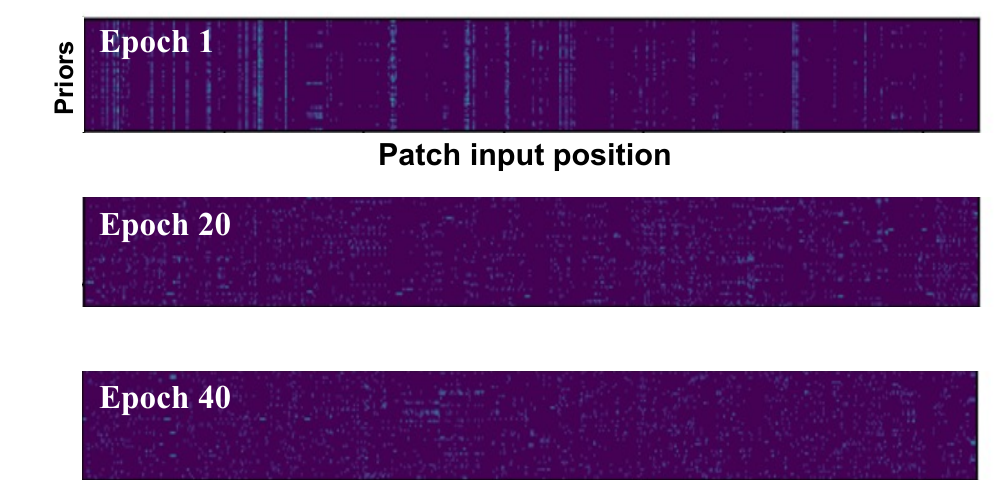}
    \caption{The Bottleneck Attention Balance Loss facilitates the selection of more diverse tokens from various input positions.}
    \label{fig:loss}
\end{figure}

\begin{figure}[!h]
    \centering
    \includegraphics[width=\linewidth]{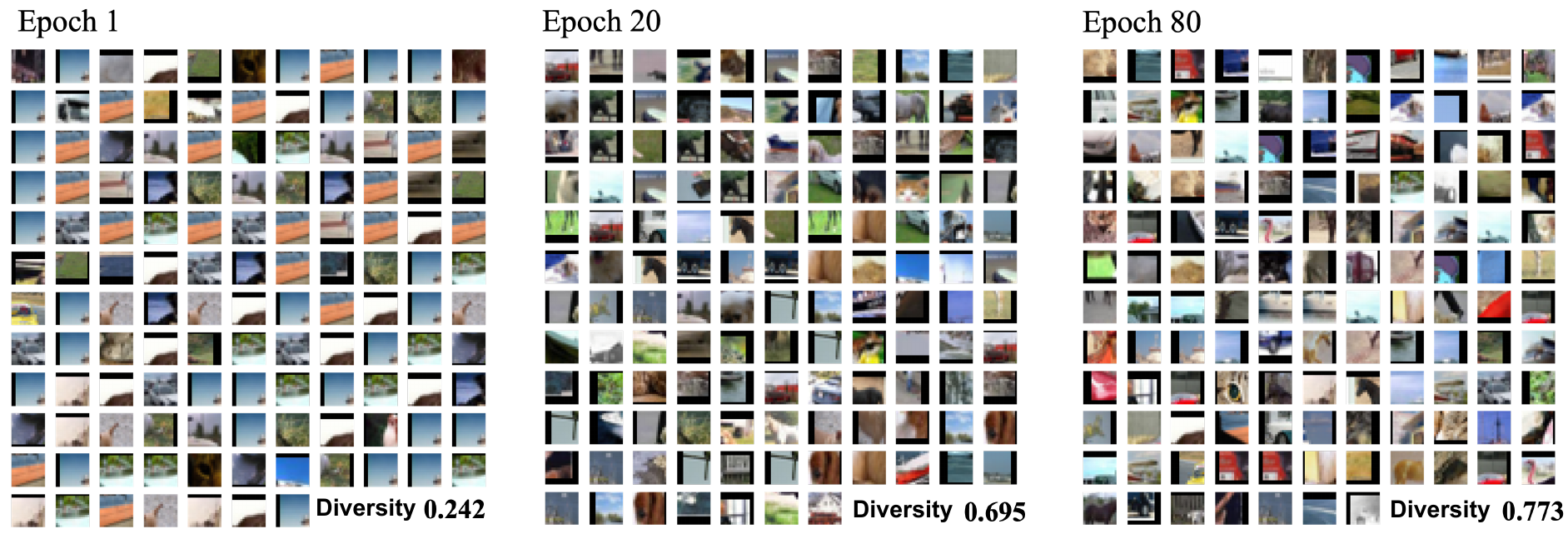}
    \caption{The corresponding patches of the selected tokens by the bottleneck attention in CIFAR-10, demonstrating an apparent increase in the diversity of selected tokens as training progresses.}
    \label{fig:spar}
\end{figure}

The Bottleneck Attention Balance Loss facilitates the learning of priors that attend to diverse sets of tokens. We demonstrate the efficacy by visualizing the bottleneck attention scores (Figure \ref{fig:loss}) and the corresponding patches of the selected tokens by the bottleneck attention (Figure \ref{fig:spar}). We employed as a metric for patch diversity the ratio of distinct tokens in all the selected tokens. 

\begin{figure}[!h]
    \centering
    \includegraphics[width=0.8\linewidth]{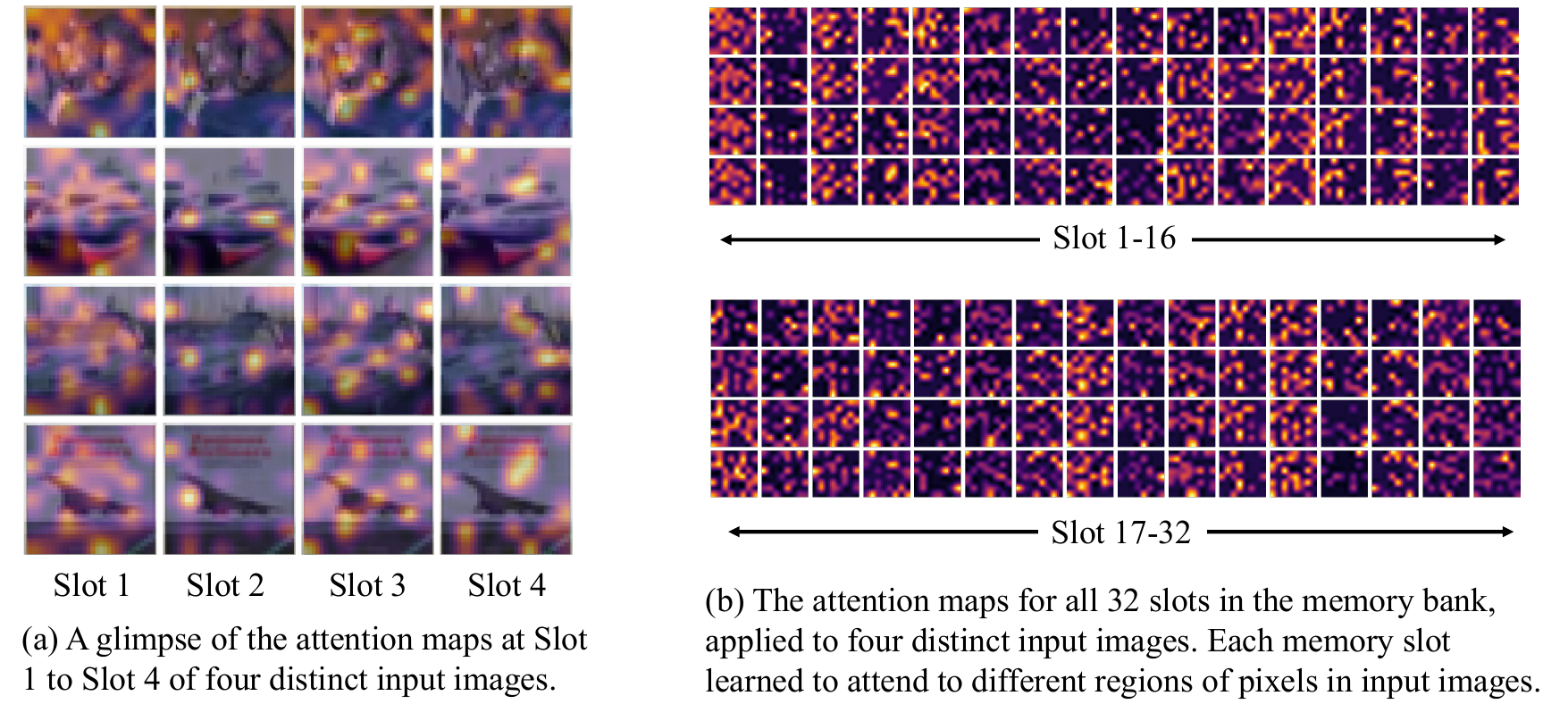}
    \caption{Prior-guided attention visualization highlights selected tokens, with each prior having learned to extract from similar patch input positions across different samples. These priors exhibit emergent specialization, focusing on specific spatial areas in an image (using the first global workspace layer of AiT-Small in CIFAR-10).}
    \label{fig:spec}
\end{figure}

We visualized the attention scores over different tokens in Figure \ref{fig:spec}. Despite employing tokens across different batch samples, each prior eventually learned to extract from similar patch input positions, demonstrating emergent spatial specialization.

\section{Hopfield networks energy}
\label{appe:hopf}

In traditional Hopfield networks, it is possible to store $N$ samples and retrieve them with partially observed or noisy patterns by updating model weights. During retrieval, these partially observed or noisy patterns converge to one of these attractors, minimizing the Hopfield energy. Unlike traditional Hopfield attractors that incorporate the implicit memory within its model parameters, AiT decouples the memory from the Hopfield network by introducing the learnable explicit memory. This memory serves the functions of both priors in the bottleneck attention and attractors in Hopfield networks. Consequently, the Hopfield network does not need to store different inner states every batch time, instead, we can reuse the learned memory bank from the bottleneck attention to update and maintain a set of attractors with the trainable linear transformation. The proposed architecture is an attractor network in the sense that, in every batch, a pattern converges to one of these attractors derived from the priors stored in the explicit memory bank. 

\begin{figure}[!h]
    \centering
    \includegraphics[width=0.9\linewidth]{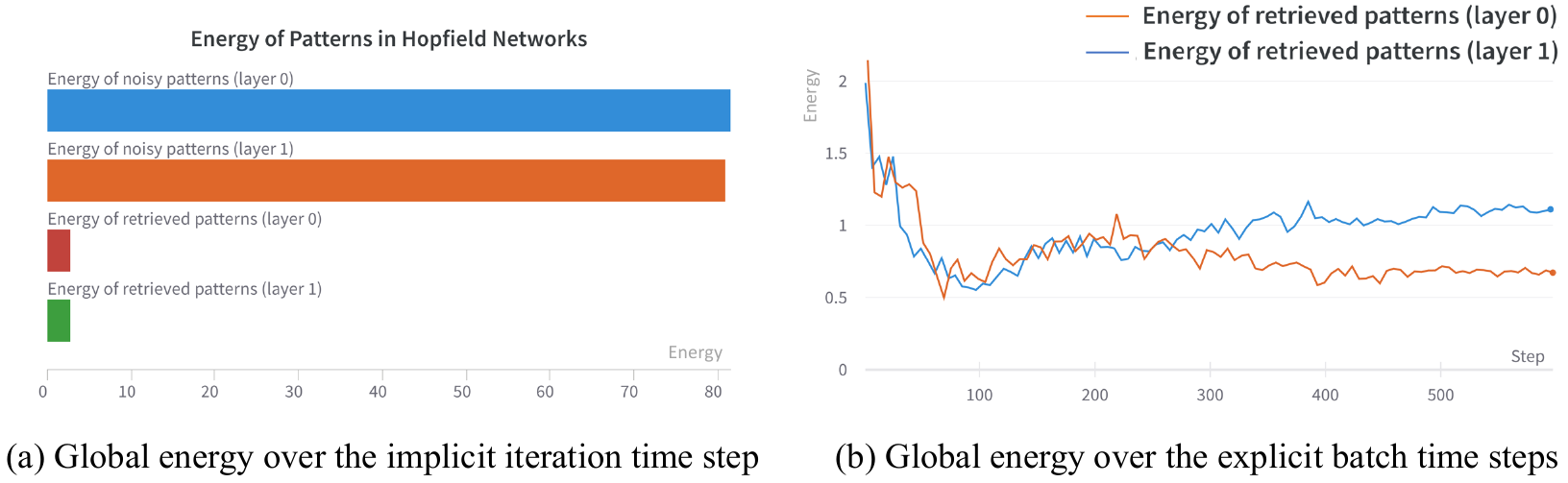}
    \caption{AiT-Small's patch representation energy for the CIFAR-10 task. The Hopfield network operates by iteratively decreasing the energy of an input state with respect to the attractors stored in its memory. This reduction in energy enables the retrieval of a representation that closely aligns with learned attractors, effectively leveraging knowledge within the associative memory. The energy is guaranteed to decrease over the iteration time step for every retrieval. Over the batch time steps, the energy generally decreases, especially during the early stages of training.}
    \vspace{10pt}
    \label{fig:hopf}
\end{figure}

Moreover, the information retrieval is based on a continuous Hopfield network, where an input state converges to a fixed attractor point within the associative memory of the Hopfield network. Usually, any input state that enters an attractor’s basin of attraction will converge to that attractor. The convergence results in a decreased state energy with respect to the stored attractors in the memory. All tokens reach their minimum at the same time and the global energy in Equation \ref{eq:ener} is guaranteed to decrease. To quantitatively measure the amount of energy reduction during the information retrieval process in the Hopfield network, we computed an input state's energy before and after it was reconstructed. A successful retrieval results in substantial reduction in the state energy (Figure \ref{fig:hopf}).

\begin{figure*}[!h]
    \centering
    \includegraphics[width=\linewidth]{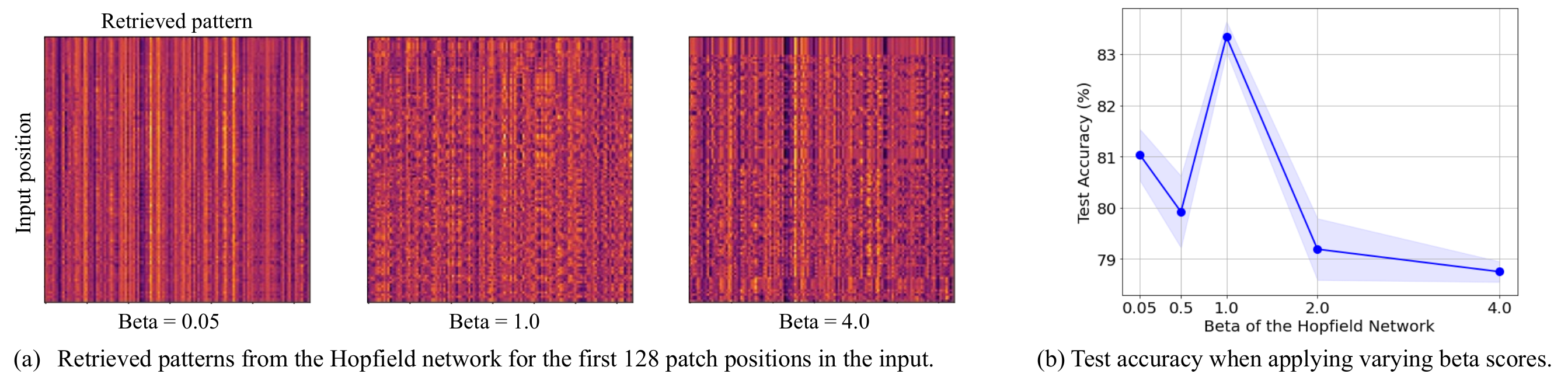}
    \caption{Varying the inverse temperature score influences the formation of the metastable states that are mixtures of patch representations. A smaller $\beta$ is more likely to generate such metastable states, while a larger $\beta$ leads to a stronger separation of different patterns. However, the results showed that a larger $\beta$ could also lead to local minima, where input patterns were reconstructed to the same pattern.}
    \label{fig:temp}
\end{figure*}

Furthermore, we investigated the effect of the inverse temperature $\beta$ on the information retrieval capability of Hopfield networks in Figure \ref{fig:temp}. We found that using an inverse temperature of 1.0 obtained the best retrieval performance based on the Hopfield networks. The results suggest that the inverse temperature parameter requires tuning to reach optimal performance. We aim to study a mechanism to adjust $\beta$ adaptively in a future study.



\end{document}